\documentclass[letterpaper, 10 pt, conference]{ieeeconf}  

\IEEEoverridecommandlockouts                              

\usepackage{graphicx,subcaption} 
\usepackage{amsmath} 

\makeatletter               
\let\NAT@parse\undefined
\makeatother
\usepackage{hyperref}
\hypersetup{colorlinks,breaklinks,
linkcolor=[rgb]{0.5,0.,0.},
citecolor=[rgb]{0.000,0.427,0.173},
urlcolor=[rgb]{0.031,0.318,0.612}}

\usepackage{amsfonts}

\DeclareMathOperator*{\argmin}{arg\,min}

\title{\LARGE \bf
Confidence-Aware Decision-Making and Control for Tool Selection*
}

\author{Ajith Anil Meera$^{1}$ and Pablo Lanillos$^{1,  2}$
\thanks{*This research is supported by the Metatool project (Grant agreement 101070940) under the EIC pathfinder program. Corresponding author: {\tt\small ajith.anilmeera@donders.ru.nl}}
\thanks{$^{1,2}$Donders Institute, Radboud University Nijmegen, The Netherlands
        }%
\thanks{$^{2}$Cajal International Center for Neuroscience, Spanish Research Council.}%
}

\begin{document}

\maketitle
\thispagestyle{empty}
\pagestyle{empty}

\begin{abstract}
Self-reflecting about our performance (e.g., how confident we are) before doing a task is essential for decision making, such as selecting the most suitable tool or choosing the best route to drive. While this form of awareness---thinking about our performance or metacognitive performance---is well-known in humans, robots still lack this cognitive ability. This reflective monitoring can enhance their embodied decision power, robustness and safety. Here, we take a step in this direction by introducing a mathematical framework that allows robots to use their control self-confidence to make better-informed decisions. We derive a mathematical closed-form expression for control confidence for dynamic systems (i.e., the posterior inverse covariance of the control action). This control confidence seamlessly integrates within an objective function for decision making, that balances the: i) performance for task completion, ii) control effort, and iii) self-confidence. To evaluate our theoretical account, we framed the decision-making within the tool selection problem, where the agent has to select the best robot arm for a particular control task. The statistical analysis of the numerical simulations with randomized 2DOF arms shows that using control confidence during tool selection improves both real task performance, and the reliability of the tool for performance under unmodelled perturbations (e.g., external forces). Furthermore, our results indicate that control confidence is an early indicator of performance and thus, it can be used as a heuristic for making decisions when computation power is restricted or decision-making is intractable. Overall, we show the advantages of using confidence-aware decision-making and control scheme for dynamic systems. 
\end{abstract}

\section{INTRODUCTION}

Humans make decisions not only based on performance but also on their confidence in decisions \cite{yeung2012metacognition}. For example, humans might prefer to use a known (confident) hammer to hit a nail, instead of a high-performance nail gun. Similarly, while driving, she might prefer a known route instead of an unknown path that might be optimal in terms of distance~\cite{bontje2023you}. Thanks to this capability to self-assess our performance (metacognitive performance)~\cite{fleming2012metacognition}, we make decisions that can handle (or adapt to) uncertainties~\cite{yeung2012metacognition} (e.g., a bent nail, road closings). 
Whilst confidence is an integral part of decision-making in humans~\cite{fleming2017self}, and can provide more decision power and robustness, it is not widely used in robotics. Decision-making in robotics usually optimizes a performance-based objective \cite{parkan1999decision}, such as the quadratic function of state and control for optimal control~\cite{anderson2007optimal} (e.g., LQR). 


The concept of confidence has recently gained relevance in robotic applications, such as autonomous driving \cite{bontje2023you}, shared control \cite{saeidi2018confidence,saeidi2015trust} and collision avoidance~\cite{fridovich2020confidence}, where they employed empirical and learning methods \cite{cao2021confidence} to measure the confidence in decisions. These methods usually evaluate human predictability confidence and not the robot's confidence in executing its own actions. 
In essence, in any decision-making problem (e.g., selecting the best action or classifying images) the idea is to be able to compute the posterior uncertainty of making a decision~\cite{moon2020confidence}---providing the true underlying distribution. In control applications, we frame this confidence as how sure the robot is about making an optimal/best action. Importantly, while in discrete decision-making problems, there are ways to calculate these confidence estimates~\cite{papadopoulos2001confidence}, computing them for continuous-time dynamical systems is less explored, as they mainly focus on the posterior distribution of predictions ~\cite{lederer2020confidence}, or do not use the control confidence \cite{kouw2023information}.


Here, we draw inspiration from the metacognitive approach to cognition~\cite{fleming2017self,hesp2021deeply}, to propose and derive an analytical expression to compute control confidence in dynamical systems. Thus, allowing the agent to use performance and confidence estimates to make better-informed decisions. To this end, we use a probabilistic approach to control, based on the Free Energy Principle (FEP)~\cite{friston2010free,lanillos2021active}, which permits the calculation of the control confidence as the posterior inverse covariance of the control action. Hence, we provide a solution in continuous time, state and action space that incorporates the robot's self-confidence within the decision-making process, taking the dynamics of the system into account. The core theoretical contributions are: 
\begin{enumerate}
    \item a closed-form expression for the control confidence,
    \item a confidence-aware objective that balances between performance, control effort and control confidence,
    \item a mathematical framework for decision-making and control using this objective.
\end{enumerate}

\begin{figure*}[hbtp!]
\centering
\vspace{.2cm}
\includegraphics[scale = 0.64]{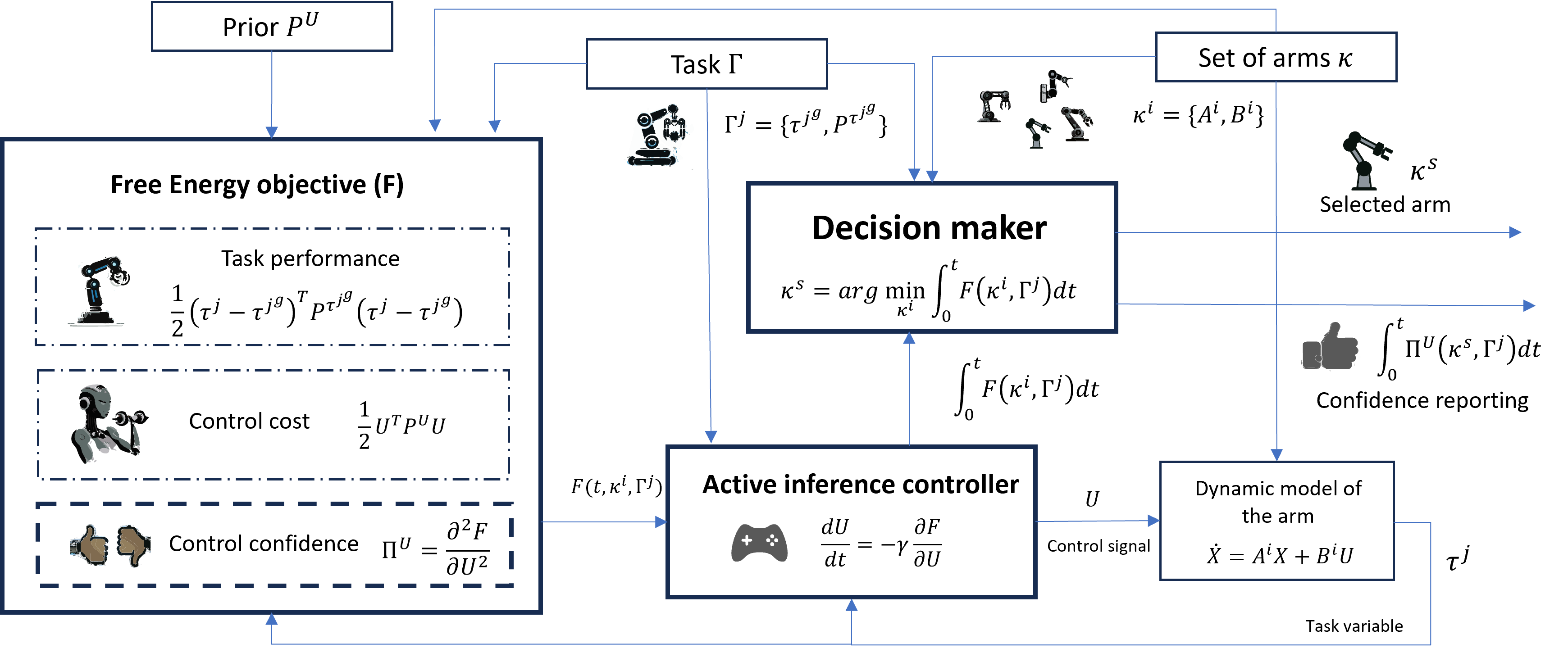}
\caption{The schematic of our proposed confidence aware decision making and control scheme. The robot makes a decision and reports its confidence. The decision making criteria simulates the controller to evaluate the free energy objective that balances task performance, control effort and control confidence. The self evaluation (performance) and self reflection on its decision (confidence) makes the robot metacognitive. The first two gradients of $F$ fully describes our controller  design (contoller and control confidence), while the integral of $F$ fully describes our decision maker. See Section \ref{sec:theory} for the mathematical details. }
\label{fig:schematic}
\end{figure*}

These theoretical contributions are supported by its evaluation of the working example of the tool selection problem~\cite{anil2023towards}, where the robot has to select the tool that best performs the task. Without loss of generality, we define the set of selectable tools as a subset of possible robot arms---See Fig.~\ref{fig:schematic}. 
Statistical evaluation of this decision-making problem shows the usefulness of including control confidence. In particular, results show that the selected tool/arm using our proposed criteria is less sensitive to perturbations (while doing the task) than tools selected using the performance-based objective. Our results also point towards the use of control confidence as an early and easy-to-compute indicator of performance, without any prediction horizon unrolling. This increases its practical relevance, especially for fast decision-making under limited computation time/power. 

\section{PROBLEM FORMULATION} \label{sec:problem_statement}
We use the tool selection problem as the driving example of confidence-aware decision-making. The tool selection problem consists of selecting a tool (defined as a robotic arm) from a set of $p$ tools $\mathbf{\kappa} = \{\kappa^1,\kappa^2,\ldots, \kappa^p\}$, such that it best completes a task using its controller, by fulfilling the desired goal conditions. We restrict the set of possible tools to those whose dynamics can be modelled using a linear state space system of the form:
\begin{equation} \label{eqn:LTI_general}
     \dot{X} = AX+BU,
\end{equation}
where $A\in \mathbb{R}^{n \times n}$ and $B \in \mathbb{R}^{n \times r}$ are the matrices defining the system dynamics, $U \in \mathbb{R}^{r \times 1}$ is the control input to the system and $X \in \mathbb{R}^{n \times 1}$ is the hidden state. Hence, every tool is fully described by its matrices $\kappa^i= \{A^i,B^i\}$. 

We consider three types of tasks $\mathbf{\Gamma}= \{\Gamma^1,\Gamma^2,\Gamma^3\}$ in terms of goal condition $\tau^g$: reach a desired i) constant goal state $X^g$ (e.g., reaching task), ii) constant goal state velocity $\dot{X}^g$ (e.g., screw tightening tasks), and iii) constant goal acceleration $\Ddot{X}^g$ (e.g., constant force tasks like lifting, pushing). Thus, the problem reduces to select the tool $\kappa^i$, for a given task $\Gamma^j$, such that the task variable $\tau$ best reaches the desired goal $\tau^g$ within a tolerance (desired goal covariance) of $\Sigma^{\tau^g} = (P^{\tau^g})^{-1}$. The equations of motion of the tool (robot arm) used in the results are derived in Appendix \ref{sec:generative_process}. In Appendix \ref{sec:generative_model}, they are linearized to a form as given in (\ref{eqn:LTI_general}).

\section{CONFIDENCE-AWARE DECISION MAKING AND CONTROL} \label{sec:theory}
This section lays down the theoretical contributions that allow the robot to use confidence within the decision-making process. 
Figure \ref{fig:schematic} shows the schematic of our proposed decision-making framework driven by the tool selection problem and the three main theoretical contributions: i) the design of a task-based probabilistic controller that minimizes the free energy objective, ii) the usage of this controller to derive a closed-form expression for control confidence, and iii) introduce a decision-making process that balances performance, control cost and control confidence. We fully describe our controller design and the control confidence using the first two gradients of free energy and our decision maker using the free energy integral.

We mathematically define control confidence---the robot's confidence in the control action being the optimal one for task completion---as the posterior inverse covariance of the control action. To derive the closed-form expression of control confidence, we employ the mathematical framework of active inference, as it provides a way to evaluate it: the posterior inverse covariance is the curvature of the free energy objective. 

\subsection{Active inference controller design}
We derive an active inference controller so that the closed-form expression for the control confidence (posterior inverse covariance of control action) can be computed. To this end, we introduce a novel form of the free energy objective from first principles aimed at task completion with high performance, minimal control effort and high confidence. This objective is minimized by the controller to select the best actions~\cite{oliver2021empirical,lanillos2021active}.

Free energy optimization is a way to approximate Bayesian inference. The goal is to estimate the Bayesian optimal value for the parameter $\alpha$ such that it best explains the observation $y$. According to Bayes rule, the posterior distribution $p(\alpha/y)$ of parameter $\alpha$, given the measurement $y$ is given by $p(\alpha/y) = p(\alpha,y)/p(y)$. Since the computation of $p(y)= \int p(y,\alpha) d\alpha$ is intractable for large search spaces of $\alpha$, variational methods use a recognition density $q(\alpha)$ to closely approximate $p(\alpha/y)$ by minimizing the Kullback–Leibler (KL) divergence between both the distributions. This procedure results in the minimization of an objective function called free energy, given by \cite{friston2010free}: 
\begin{equation} \label{eqn:FE_probability}
    F = -\int q(\alpha)\ln{p(y,\alpha)} d\alpha + \int q(\alpha) \ln{q(\alpha)} d \alpha.
\end{equation}
FEP is based on the idea that the Bayesian optimal estimate of $\alpha$ is the one that minimises the free energy objective $F$. The active inference controller (based on FEP) follows the gradient of $F$ for the controller design.

\subsubsection{Task-based active inference}
We consider the problem of evaluating the control action $U$, to perform the task $\Gamma^j$, by controlling the task variable $\tau^j$ to reach the goal $\tau^{j^g}$, within a desired level of uncertainty or prior covariance $\Sigma^{\tau}=(P^{\tau^g})^{-1}$. We proceed towards modelling the distributions inside (\ref{eqn:FE_probability}), to compute $F$ for our problem definition in Section \ref{sec:problem_statement}. We model the posterior distribution of control actions using a Gaussian distribution. So, the recognition density becomes $q(U)=\mathcal{N}(U:\mu^U,(\Pi^U)^{-1})$.  The notation $P$ is used for the prior precision (or inverse covariance) and $\Pi$ is used for the posterior precision. We enable the robot to fuse its prior information into the scheme by modelling the prior distribution over control actions as a Gaussian distribution on $U$, written as $p(U)=\mathcal{N}(U:\eta^U,(P^U)^{-1})$.  We assume that we can directly measure the task variable ($y = \tau^j$), and choose the control action as the unknown parameter to be computed ($\alpha=U$). We model the posterior distribution of the task variable under the robot's action  using a Gaussian distribution, given by $p(\tau^j/U)=\mathcal{N}(\tau^j:\tau^{j^g},(P^{\tau^j})^{-1})$.  The free energy in (\ref{eqn:FE_probability}) is simplified using the distributions, and the constants are dropped to get (See Appendix \ref{app:FE_derivation} for the detailed derivation):
\begin{equation} \label{eqn:FE_components}
\begin{split}
F = & \ \underbrace{ \frac{1}{2}(\tau^j -\tau^{j^g})^T P^{\tau^{j^g}} (\tau^j -\tau^{j^g})  }_{\text{performance}}+ \underbrace{ \frac{1}{2}U^T P^{U} U}_{\text{control cost}}  - \underbrace{ \frac{1}{2} \ln{|\Pi^U|}}_{\text{confidence}}.
\end{split}
\end{equation}
The resulting free energy objective can be seen as a sum of three terms: i) performance measure (the prior precision weighted deviation of the task variable from the goal), ii) control cost (the prior precision weighted control effort), and iii) confidence measure(the level of confidence in the chosen control action). When $\eta^U =0$, minimizing $F$ implies, maximizing performance, minimizing control effort and maximizing confidence. This objective can be used to design the controller. Our agent optimizes $F$ to choose the best control actions via gradient descent on free energy: 
\begin{equation} \label{eqn:AIC_gradient_des}
    \frac{dU}{dt} = -\gamma \frac{\partial F}{\partial U},
\end{equation}
where $\gamma$ is the learning rate. 


\subsubsection{Robotic arm control}
Here we derive the task-based active inference controller for a robot arm to reach the task goal in joint space. The free energy of an agent that wants to achieve a particular goal angular position $\theta^g$, goal angular velocity $\dot{\theta}^g$ and goal angular acceleration $\Ddot{\theta}^g$, with goal precisions (inverse covariance $P^{\theta^g}$, $P^{\dot{\theta}^g}$ and $P^{\Ddot{\theta}^g}$), and control priors ($\eta^U$, $P^U$)  is:
\begin{equation} \label{eqn:free_energy}
\begin{split}
    F  = & \frac{1}{2} \Big[ (\theta - \theta^g)^T P^{\theta^g}(\theta - \theta^g) +  (\Dot{\theta} - \Dot{\theta}^g)^T P^{\Dot{\theta}^g}(\Dot{\theta} - \Dot{\theta}^g) \\
    & +  (\Ddot{\theta} - \Ddot{\theta}^g)^T P^{\Ddot{\theta}^g} (\Ddot{\theta} - \Ddot{\theta}^g) -  \ln{|\Pi^U|} \\
    & +  (U -\eta^U)^T P^U (U-\eta^U)   \Big].   
\end{split}
\end{equation}

The core tenet of this controller is that taking control action $U$ that minimises $F$, the states of the system ($\theta, \ \dot{\theta}$ and $\Ddot{\theta}$) move close to their respective goals. When the control prior is set to $\mu^U = O$, the controller tries to minimize the control effort (or magnitude) of $U$, where $P^U$ modulates the weightage of the control effort within $F$. $F$ is similar to the objective $J$ \footnote{$J = \int_0^\infty (x^TQx+u^TRu)dt$ where $Q$ and $R$ are weights.} from optimal control \cite{anderson2007optimal}, with an additional term of $\ln{|\Pi^U|}$, which describes the control confidence, i.e., the confidence of the agent on the chosen control action to take the system to the goal (task completion) optimally.

To compute the control actions we differentiate (\ref{eqn:free_energy}) and substite it in (\ref{eqn:AIC_gradient_des}). Thus, giving the update rule for the controller:
\begin{equation} \label{eqn:U_dot_deriv}
\begin{split}
    \Dot{U}  = -& \frac{\partial \theta}{\partial U}^T  P^{\theta^g} (\theta - \theta^g)  -  \frac{\partial  \Dot{\theta}}{\partial U}^T P^{\Dot{\theta}^g}  (\Dot{\theta} - \Dot{\theta}^g)       \\ 
     - & \frac{\partial  \Ddot{\theta}}{\partial U}^T P^{\Ddot{\theta}^g} (\Ddot{\theta} - \Ddot{\theta}^g) - P^U (U - \eta^U).
\end{split}
\end{equation}
The update equations of this controller to perform the three tasks are derived in Appendix \ref{sec:task_controller}.

\subsection{Control confidence}
Inspired from the dynamics expectation maximization algorithm \cite{anil2021dynamic,friston2008variational}, we derive a closed-form solution for the optimal posterior precision (inverse covariance) of control action as the curvature of $F$ (derivation in Appendix \ref{app:control_conf}):
\begin{equation} \label{eqn:control_conf}
    \Pi^U = \frac{\partial^2 F}{\partial U^2}.
\end{equation}
The precision of control action indicates the degree of certainty that the controller has over its control action. Technically, it is the inverse covariance of the control action. Higher $\Pi^U$ indicates tighter bounds (or uncertainty) on the control $U$, which intuitively means that the controller is confident about the control action, while a lower $\Pi^U$ indicates a larger uncertainty over the control action, and lower confidence in control. In Section \ref{sec:control_perturbations}, we will show that a higher $\Pi^U$ indicates a higher stability in tool performance under perturbations on the control signal $U$. 

\begin{figure*}[]
\vspace{.1cm}
    \centering
    \begin{subfigure}[b]{0.29\textwidth}
		\includegraphics[width=\textwidth]{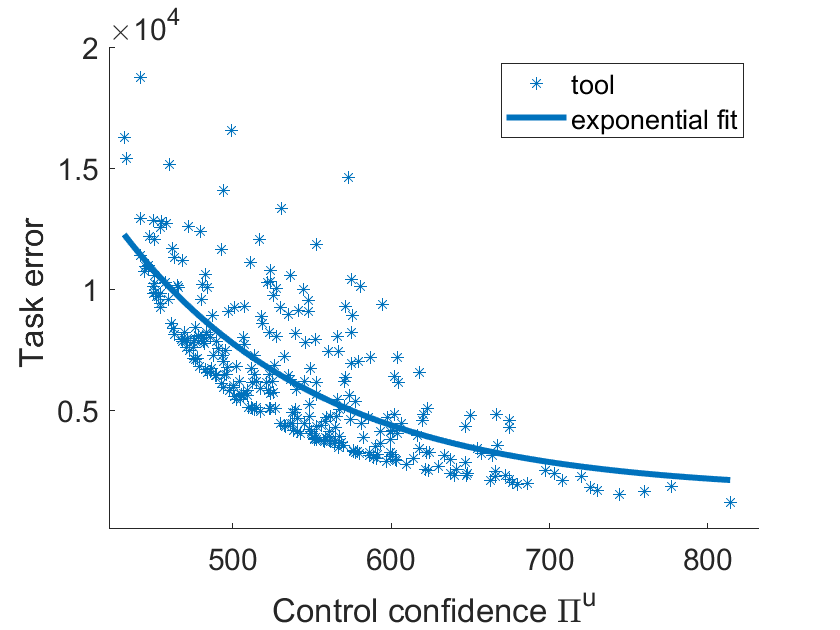}
		\caption{}
    \end{subfigure} 
    \begin{subfigure}[b]{0.29\textwidth}
		\includegraphics[width=\textwidth]{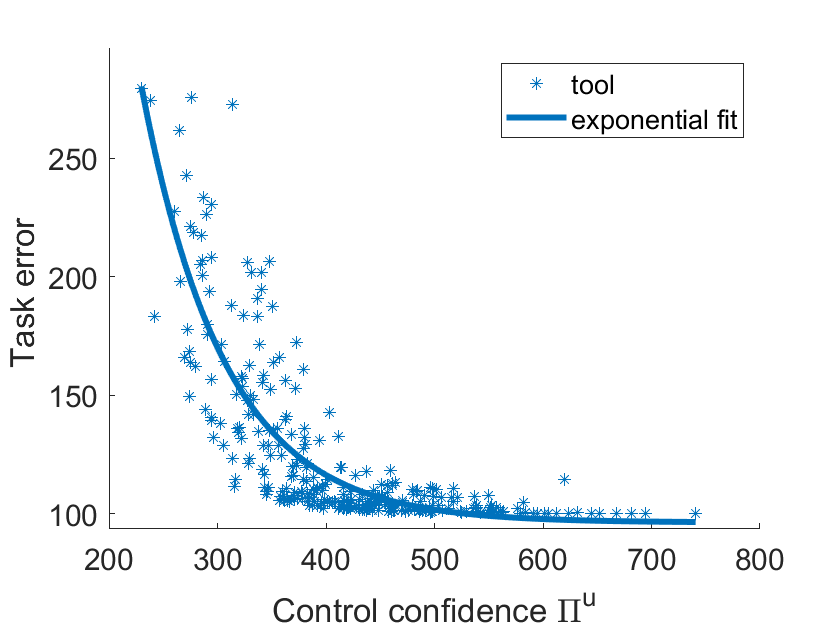}
		\caption{}
    \end{subfigure}   
    \begin{subfigure}[b]{0.29\textwidth}
		\includegraphics[width=\textwidth]{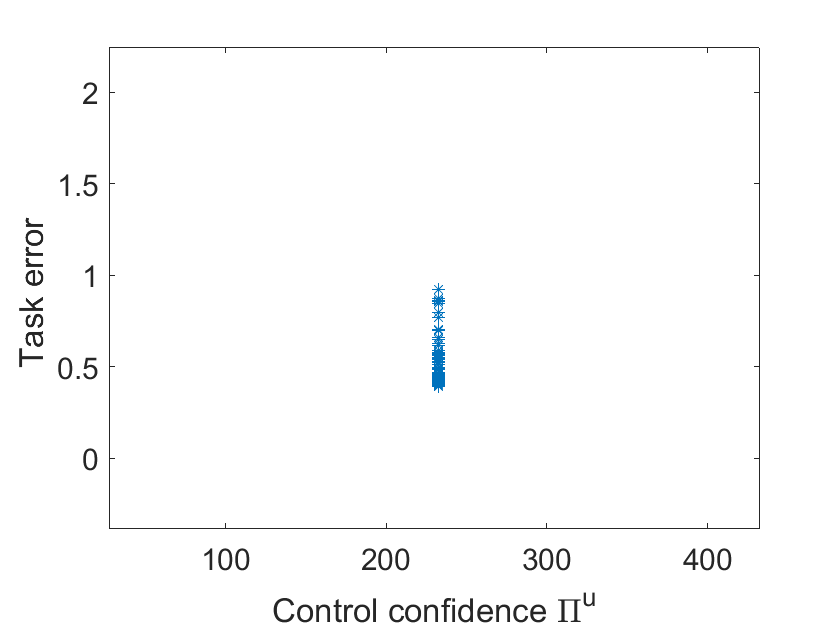}
		\caption{}
    \end{subfigure}   
    \caption{Task error vs control confidence for 300 randomly selected tools (with exponential curve fit) for three tasks a) position task. b) velocity task and c) acceleration task. On average, the tools with higher control confidence performs better at the task than the tools with lower control confidence, for task 1 and 2. However, for task 3, the controller has the same confidence on all tools, which is in line with similar task errors for all tools. }\label{fig:Conf_taskerror_task}
\end{figure*}

The closed form expression for the control confidence in (\ref{eqn:control_conf}), for position task can be derived as (see Appendix \ref{sec:task_controller} for full derivation):
\begin{equation}
    \Pi^U  =  \   (A^{-1} (\frac{\partial  \Dot{X}}{\partial U}  - B))^T P^{X^g} (A^{-1} (\frac{\partial  \Dot{X}}{\partial U}  - B))  +  P^{\Ddot{\theta}^g} + P^U, 
\end{equation}
The control confidence of the tool $\Pi^U$ depends on i) the dynamics of the tool ($A$, $B$), ii) on the prior precisions of the goal ($P^{X^g}$, $P^{\Ddot{\theta}^g}$), and iii) the agent's prior control confidence ($P^U$). Since $\Pi^U $ is independent of $X$ and $U$, and is dependent on constants representing the robot arm and the priors of the agent, $\Pi^U$ can be precomputed, and hence remains a constant while in operation. Therefore, the confidence in tool use can be evaluated before the tool use, increasing the practical relevance of this expression. 

Similarly, we can derive the control confidence for velocity and acceleration control as:
\begin{equation}
\begin{split}
 \Pi^U = & \  ([    I_{2 } \ O_{2 }]A^{-1}B ) ^T P^{\Dot{\theta}^g}  ([    I_{2 } \ O_{2 }]A^{-1}B ) + P^{\Ddot{\theta}^g} + P^U, \\
    \Pi^U = & \  ([I_2 \ O_2] A^2 (A^{-1}(\frac{\partial  \Dot{X}}{\partial U}  
 - B) + A B)^T P^{\Ddot{\theta}^g}  \\ &\  ([I_2 \ O_2] A^2 (A^{-1}(\frac{\partial  \Dot{X}}{\partial U} 
 - B) + A B)     +P^U.
\end{split}
\end{equation}

\subsection{Confidence-aware decision making}
This section introduces a confidence aware objective function that can be used for high level robot decision making. Again, we use tool selection as the driving example -- selecting the best tool from a set of tools. Conventional methods follow a performance based objective where the error to goal and the control cost is penalized within the objective \cite{anderson2007optimal}. The main contribution of this work is the introduction of a control confidence term into this decision making criteria, enabling the agent to make high level decisions by considering their confidence in each decisions made. We propose the use of a high level decision making criteria that uses the performance measure, control cost and control confidence of a low level controller, to make high level decisions about the tool selection. The tool selection problem involves the selection of a tool (robot arm) $\kappa^i$ from a set of tools $\mathbf{\kappa} = \{\kappa^1,\kappa^2,..,\kappa^p\}$, that best performs the task $\Gamma^j$. From Section \ref{sec:generative_model}, we can fully describe a robot arm as a tool using its matrices $\kappa^i = \{ 
A^i, B^i \}$. We propose a free energy based tool selection criteria to balance between the performance, control cost and control confidence as:
\begin{equation}
    \kappa^s = \argmin_{{\kappa^i}} \int_{0}^{t} F(\kappa^i,\Gamma^j)dt.
\end{equation}
The chosen tool maximises the task performance with minimal control effort and maximum confidence in action, leading to the task completion. In addition to task performance, the agent is now aware of its self confidence in actions while using the tool, making the decision making process metacognitive. The next section will deal with demonstrating the usefulness of this criteria in simulation.

\section{RESULTS}
We evaluate how control confidence affects decision-making under the tool selection problem. Results show the following:
\begin{enumerate}
    \item Sec \ref{sec:control_conf_vs_perf} shows that 
    control confidence is correlated with task performance. This means that it may be used as an early indicator of performance, for instance, to use it as a heuristic  for tool comparison.
    \item Sec \ref{sec:control_perturbations} shows that control confidence indicates the reliability of a tool for its task performance under a constant control perturbation.
    \item Sec \ref{sec:objective_selection_comparision} shows how different objective functions impact the decision making (i.e, selects different tools).
    \item Sec \ref{sec:benchmarking}  statistically shows the advantage of using $F$ over $J$ for decision making   -- $F$ chooses better tools.   
\end{enumerate}

\subsection{Implementation details}
All tools are considered to be 2DOF arms. The dynamic system used for the robot arm follows the derivations from Appendix \ref{sec:generative_process} and \ref{sec:generative_model}. The arms are defined by mass and length parameters $\{m_1,m_2,l_1, l_2 \}$, which are randomly selected from the parameter domain $[0.1,0.6]$. $ode45$ in MATLAB was used to solve the differential equations of the controller and the dynamics for 5 seconds with a sampling time of $dt=0.05$ with initial conditions of states as $X_0 = [-\frac{\pi}{2} \ -\frac{\pi}{2} \ 0 \ 0]^T$ and of torque as $T_0=[0.1 \ 0.1]^T$. Three tasks are considered: i) task 1 (position task) with goal position $\theta^g = [\frac{\pi}{3} \ \frac{\pi}{6}]^T$ and, zero goal velocity and acceleration, ii) task 2 (velocity task) with goal velocity $\dot{\theta}^g = [0 \ 2]^T$ and zero acceleration, and iii) task 3 (acceleration task) with goal acceleration $\ddot{\theta}^g = [0 \ 0.1]^T$. The goal precisions are set as $P^{\theta^g} = diag(100,100)$ for task 1, $P^{\dot{\theta}^g} = diag(50,50)$, $P^{\ddot{\theta}^g} = diag(1,1)$ and $P^u = diag(1,1)$ for task 1 and 2 and, $P^{\ddot{\theta}^g} = diag(10,10)$ and $P^u = diag(0,0)$ for task 3. The performance of each tool on a given task is evaluated using (weighted) task error as the metric. For example, the task error for task 1 is $\sum_{t} (\theta - \theta^g)^T P^{\theta^g}  (\theta - \theta^g)$. Higher the task error, lower the tool performance on the given task.

\begin{figure}[!hbtp]
\centering
\includegraphics[scale = 0.3]{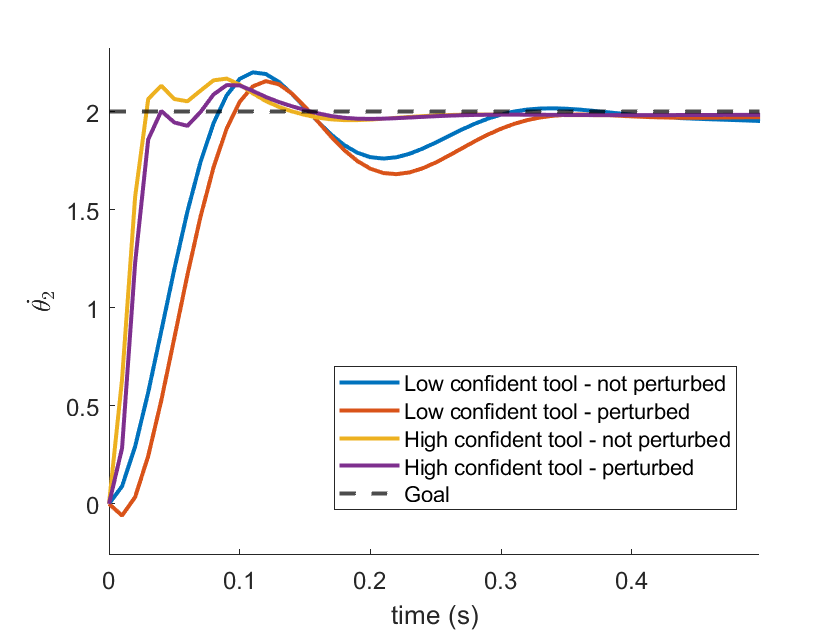}
\caption{High confident tool (yellow and purple) settles near the goal $\theta_2^g$ (in dotted black) faster than the low confident tool (red and blue). The low-confident tool shows more variability in performance than the high-confident tool under a constant control perturbation of $\Delta T = -0.8Nm$ on the joints. }
\label{fig:conf_perf_state}
\end{figure}

\begin{figure}[!hbtp]
\centering
\includegraphics[scale = 0.3]{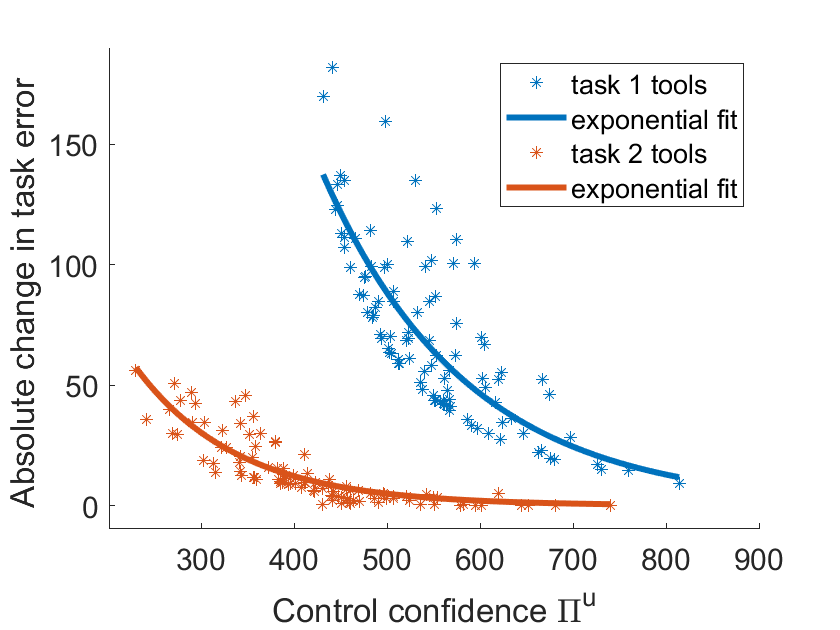}
\caption{Higher the control confidence on the tool, lower the absolute change in task error, under a control perturbation. This implies that the tool performance is less influenced by control perturbations for tools with higher $\Pi^u$. Therefore, the control confidence represents the reliability of the tool for task performance, under control perturbations. }
\label{fig:task12_changeerror_conf}
\end{figure}

\subsection{Control confidence and tool performance} \label{sec:control_conf_vs_perf}
This section aims to establish the correlation between control confidence and tool performance in simulation. 300 random tools were sampled and the controller was evaluated for three tasks. Fig. \ref{fig:Conf_taskerror_task} reports the clear correlation between the control confidence and the task error for tasks 1 and 2. On average, the tools with high control confidence performs better for tasks 1 and 2. However, for task 3, the controller is equally confident about all tools, which is in line with the similarity in their performance for all tools. In conclusion, control confidence can be used as an early indicator for tool performance. 

\begin{figure}[!hbtp]
\centering
\includegraphics[scale = 0.3]{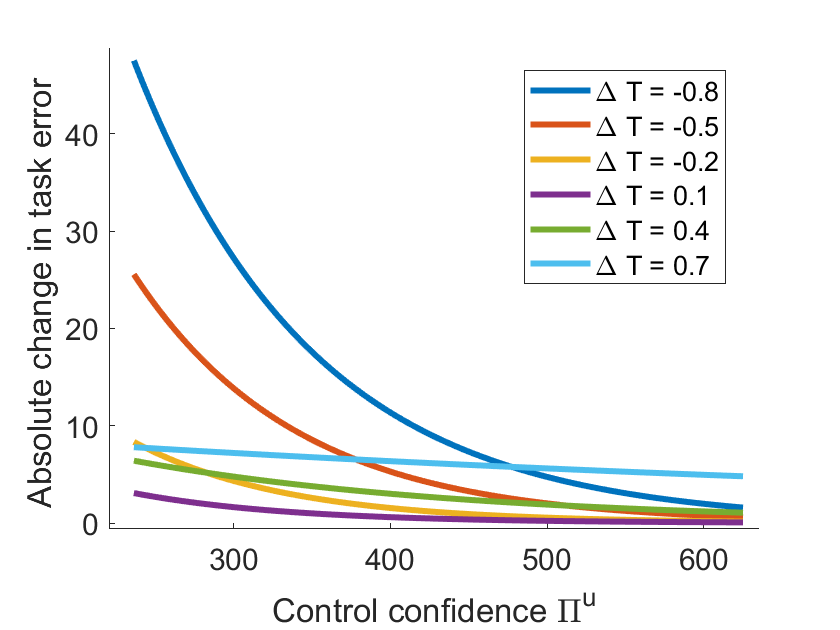}
\caption{Tools with low control confidence show high variability in task performance when subjected to different levels of constant control perturbations during the velocity task. Tools with high control confidence show relatively low variability.}
\label{fig:delta_T_conf}
\end{figure}

\subsection{Control confidence and control perturbations } \label{sec:control_perturbations}
This section aims to show that the performance of tools with high control confidence are less effected by control perturbations on the tool. Fig. \ref{fig:conf_perf_state} shows that the low confident tool shows high variability in performance than the high confident tool (under a constant control perturbation of $\Delta T = -0.8Nm$ on the joints). Moreover, the high confident tool reaches the goal faster than the low confident tool, highlighting the significance of the confidence. 100 random tools were sampled and the change in task error when the tool is subjected to a constant control perturbation of $\Delta T = -0.8$ were analysed for tasks 1 and 2. Fig. \ref{fig:task12_changeerror_conf} shows that the tools with higher control confidence show less changes in task performance under a control perturbation, whereas the tools with lower control confidence show a higher variability in task performance. Fig. \ref{fig:delta_T_conf} shows that the trend remains the same for a range of perturbations, with higher variability when the perturbation resists the controller (perturbing torque is negative and opposes the controller torque). In conclusion, control confidence can be used as a measure of variability in task performance under control perturbations.



\begin{figure*}[htb]
\vspace{.1cm}
\centering
\includegraphics[scale = .4]{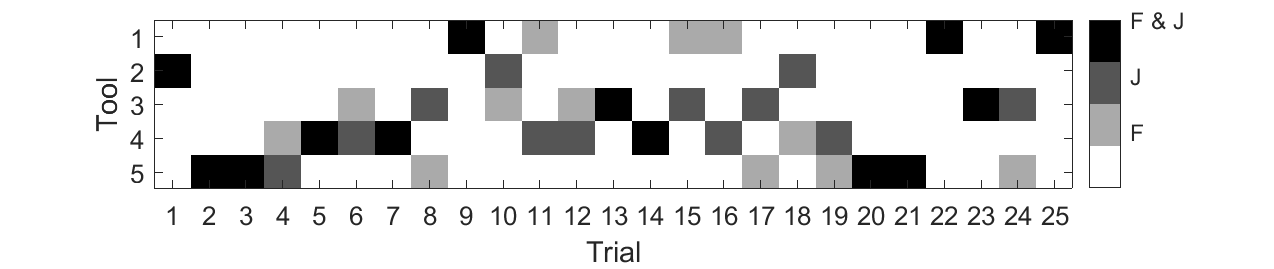}
\caption{Selected tool/arm depending on the objective function used for decision-making. The experiment uses 25 trials, with tools selected from a randomized set of 5 tools. Tool not selected (white), tool selected by $F$ (light-grey), tool selected by $J$ (dark-grey), tool selected by both $F$ and $J$ (black). The presence of grey blocks shows that both objectives can produce different decision outcomes, i.e., different tool selections.}
\label{fig:tool_selection_F_J}
\end{figure*}

\subsection{Tool selection criteria using control confidence} \label{sec:objective_selection_comparision}
We show that our tool selection criterion produces different decision-making outcomes than just optimizing for performance, i.e., it selects different tools. While the objective function (usually) used in control engineering ($J$) balances between performance and control effort, our tool selection objective ($\Bar{F}$) balances between performance, control effort and control confidence. Selecting a tool that the agent is confident about is an important aspect of tool selection. Fig. \ref{fig:tool_selection_F_J} shows the tools selected among the randomly generated toolset of 5 tools each for 25 trials. The presence of grey blocks indicates that both $F$ and $J$ do not select the same tool. Therefore, control confidence plays a significant role in selecting a different tool than the one selected through pure performance.

\subsection{Benchmarking tool selection} \label{sec:benchmarking}
\begin{figure}[!hbtp]
\centering
\includegraphics[scale = 0.4]{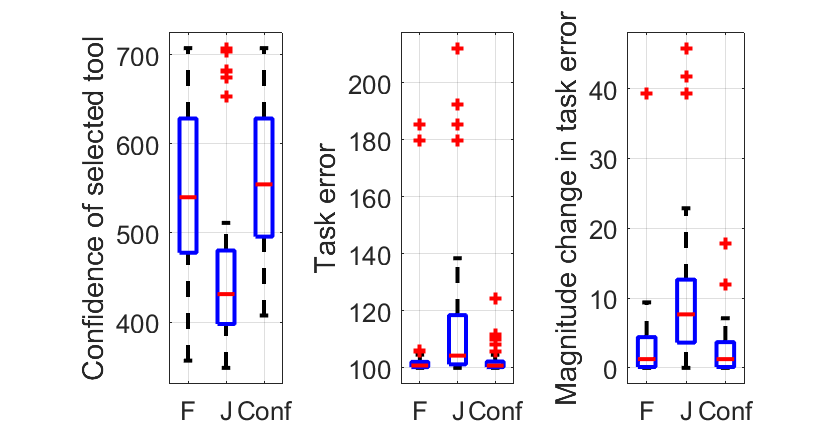}
\caption{The statistical analysis of our confidence-aware decision-making framework. We evaluate confidence, task error and the variability in task performance under a constant control perturbation (external joint torque). The $F$ objective (on average)  selects tools with a high control confidence (leftmost figure), and low variability in task performance under control perturbations (rightmost figure). $F$ has the lowest average task error (middle figure) while maintaining the reliability of pure confidence-based objective.}
\label{fig:tool_quality}
\end{figure}

We statistically show the advantage of our tool selection objective over the one used in control engineering. This is done by showing that the tool selected using $F$ has a superior tolerance to control perturbations than the tool selected using $J$. Fig. \ref{fig:tool_quality} shows the benchmarking for tool selection for task 2 using three objectives: $F$, $J$ and a pure confidence-based tool selection. 10 tools each were randomly selected for 50 trials and the performance of the tools selected by all three objectives were analysed. The tools selected by $J$ showed lower control confidence and had a higher sensitivity to a constant control perturbation. The confidence-based tool selection showed similar confidence, task error and sensitivity to that of $F$. Therefore, control confidence can be used as an early predictor for tool selection, without simulating the controller performance on the tool.

\section{CONCLUSION}
Metacognitive performance evaluation (evaluation of one's self-confidence in their decisions) is a fundamental cognitive feature that differentiates decision-making in humans and robots. We drew inspiration from this to introduce a mathematical framework for confidence-aware decision-making and control for tool selection. We derived a closed-form expression for the control confidence, which can be precomputed by the robot using the known system dynamics. Using this confidence, the proposed objective function that the decision-making employs seamlessly combines task performance, control cost and control confidence. Through both qualitative and statistical numerical experiments, we showed the advantages and properties of confidence-aware decisions in control. Our decision-making approach chose tools that presented less variability in task performance under control perturbations (external force) than the tools selected by the performance-based objective from optimal control. Relevantly, we showed that the control confidence can be precomputed and that it is correlated with task performance, suggesting its use as an early indicator/heuristic of tool performance for a given task. Future research will focus on applying the proposed theoretical framework to realistic tasks, such as industrial applications. Besides, we are investigating how to online adapt the robot's confidence during task execution.




\bibliographystyle{IEEEtran}
\footnotesize
\bibliography{main}

\appendix


\subsection{Tool Dynamics (generative process)} \label{sec:generative_process}
\begin{figure}[!ht]
    \centering
    \includegraphics[scale = .45,trim={0cm 0 0cm 0},clip]{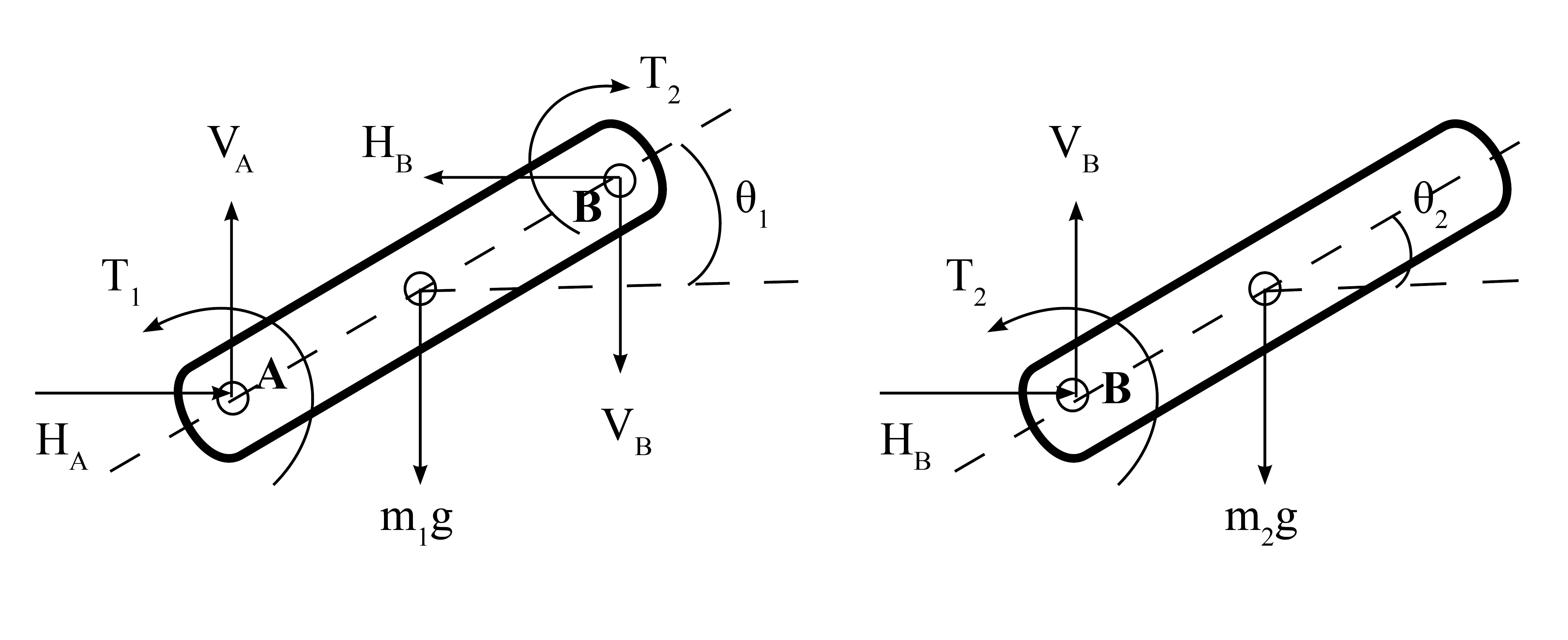}
    \caption{Free body diagram of the links of a 2DOF robot arm.}
     \label{fig:FBD}
\end{figure}

This section details the mathematical model that describes the dynamics of a 2DOF robot arm (in the vertical plane), which will be used as the tool. The free body diagram of the links with input torques $T_1$  and $T_2$ applied to joints $A$ and $B$ is shown in Fig. \ref{fig:FBD}.
The force balance equation on link 1 in $x$ and $y$ directions is written as:
\begin{equation}
    m_1 \ddot{y_1} = V_A - V_B -m_1 g , \ m_1 \ddot{x_1} =  H_A - H_B .
\end{equation}
The moment balance equation at joint $A$ can be written as:
\begin{equation} \label{eqn:link1_moment}
\begin{split}
    \frac{1}{3} m_1 l_1^2 \ddot{\theta_1} = T_1 - T_2 & + H_B l_1 \sin{\theta_1} -V_B l_1 \cos{\theta_1}  \\ 
     & - \frac{1}{2} m_1 g l_1 \cos{\theta_1}.
\end{split}
\end{equation}
Similarly, the force balance and moment balance equations for link 2 can be written as:
\begin{equation} \label{eqn:link2_force}
    m_2 \Ddot{y_2}  =  V_B - m_2 g, \ m_2 \Ddot{x_2} = H_B ,
\end{equation}
\begin{equation} \label{eqn:link2_moment}
\frac{1}{3} m_2 l_2^2 \Ddot{\theta_2}  =  T_2 - \frac{1}{2} m_2 g l_2 \cos{\theta_2}.
\end{equation}    
The constraint equations for the arm to maintain its structural integrity is written as:
\begin{equation}
x_2  = l_1 \cos{\theta_1} + \frac{1}{2} l_2 \cos{\theta_2}, \ y_2  = l_1 \sin{\theta_1} + \frac{1}{2} l_2 \sin{\theta_2}.
\end{equation}
Differentiating it twice yields the relationship between angular and linear accelerations as:
\begin{equation} \label{eqn:acc_ang_lin}
\begin{split}
    \Ddot{x}_2 & = -l_1 \Dot{\theta}_1^2 \cos{\theta_1}  - l_1 \Ddot{\theta}_1 \sin{\theta_1}  -\frac{l_2}{2} \Dot{\theta}_2^2 \cos{\theta_2}  - \frac{l_2}{2} \Ddot{\theta}_2 \sin{\theta_2}     ,\\
    \Ddot{y}_2 & = l_1 \Ddot{\theta}_1 \cos{\theta_1}  - l_1 \Dot{\theta}_1^2 \sin{\theta_1}  + \frac{l_2}{2} \Ddot{\theta}_2 \cos{\theta_2}  - \frac{l_2}{2} \Dot{\theta}_2^2\sin{\theta_2} .
\end{split}
\end{equation}
Substituting (\ref{eqn:acc_ang_lin}) in (\ref{eqn:link2_force}) simplifies $V_B$ and $H_B$, which upon substitution in (\ref{eqn:link1_moment}) gives:
\begin{equation} \label{eqn:DE_theta_2}
\begin{split}
    & \Big( \frac{m_1}{3} + m_2 \Big) l_1^2 \Ddot{\theta}_1  + 
     \frac{1}{2}m_2 l_1 l_2 \Ddot{\theta}_2 \cos{(\theta_1-\theta_2)} \\ 
     & + m_2 l_1^2 \Dot{\theta}_1^2 \sin{(\theta_1 - \theta_2)} + \frac{1}{2} m_2 l_1 l_2 \Dot{\theta}_2^2 \sin{(\theta_1 - \theta_2)} \\
     & + T_2 -T_1 + \Big( \frac{m_1}{2} + m_2 \Big) g l_1 \cos{\theta_1} = 0    .
\end{split}
\end{equation}
Equations (\ref{eqn:DE_theta_2}) and (\ref{eqn:link2_moment}) are the nonlinear differential equations that represent the dynamics of the 2 DOF robot arm, given the torques $T_1$ and $T_2$. They represent the generative process of the system.

\subsection{Generative model} \label{sec:generative_model}
This section derives the simplified model of the plant (from Section \ref{sec:generative_process}) that the robot can use for the controller design and the decision making. This is done by linearizing (\ref{eqn:DE_theta_2}) and (\ref{eqn:link2_moment}) using the Jacobians around the equilibrium point to derive its linear state space form. Equation (\ref{eqn:link2_moment}) can be rearranged as:
\begin{equation} \label{eqn:theta_2_ddot}
    \Ddot{\theta}_2  =  \frac{3}{m_2 l_2^2} \Big( T_2 - \frac{1}{2} m_2 g l_2 \cos{\theta_2} \Big),
\end{equation}
which upon substitution in (\ref{eqn:DE_theta_2}) results in:
\begin{equation} \label{eqn:theta_1_ddot}
\begin{split}
    \Ddot{\theta}_1  = & -\Big( ( \frac{m_1}{3}+m_2)l_1^2 \Big) ^{-1} \Big[m_2 l_1^2 \Dot{\theta}_1^2 \sin{(\theta_1 - \theta_2)} + T_2 -T_1 \\ 
     & + \frac{3 l_1}{2 l_2}    \cos{(\theta_1-\theta_2)} \Big( T_2 - \frac{1}{2} m_2 g l_2 \cos{\theta_2} \Big)  \\
     &  + \frac{1}{2} m_2 l_1 l_2 \Dot{\theta}_2^2 \sin{(\theta_1 - \theta_2)}   + \Big( \frac{m_1}{2} + m_2 \Big) g l_1 \cos{\theta_1}  \Big]
\end{split}
\end{equation}
Using (\ref{eqn:theta_1_ddot}) and (\ref{eqn:theta_2_ddot}), a combined state space equation of the form $\dot{X} = f(X,U)$ can be written, where $X = [\theta_1 \ \theta_2 \ \Dot{\theta}_1 \ \Dot{\theta}_2]^T$ is the combined vector of joint angles and velocities, $U = [T_1 \ T_2]^T$ is the torque vector, and $f(X,U)$ is the combined function mapping. Here $f_1(X,U)=\Dot{\theta}_1$, $f_2(X,U)=\Dot{\theta}_2$, and $f_3(X,U)$ and $f_4(X,U)$ are on the right hand side of (\ref{eqn:theta_1_ddot}) and (\ref{eqn:theta_2_ddot}) respectively. Since $f$ is nonlinear, we linearize it around the equilibrium point $X^e = [-\frac{\pi}{2} \ -\frac{\pi}{2} \ 0 \ 0]^T$ as:
\begin{equation}
    \dot{X} = f(X^e,U^e) + A (X - X^e) +B (U-U^e), 
\end{equation}
where $A = \frac{\partial f}{\partial X} \Big|_{X^e,U^e}$ and $B= \frac{\partial f}{\partial U} \Big|_{X^e,U^e}$ are the Jacobians at the equilibrium point. At equilibrium point, $f(X^e,U^e)=0$. Upon simplification, the linear state space equation of a 2DOF robot arm can be written in the form of (\ref{eqn:LTI_general}) as: 
\begin{equation} \label{eqn:LTI}
    \dot{X} = A (X - X^e) +B U, 
\end{equation}
where $A$ and $B$ are: 

$A= \begin{bmatrix}
    0 & 0& 1 & 0 \\ 0 & 0 & 0 & 1 \\
    -\frac{3(m_1+m_2)g}{l_1(m_1+3 m_2)} & \frac{9 m_2 g}{4 l_1 (m_1+3 m_2)} & 0 & 0 \\
    0 & -\frac{3g}{2 l_2} & 0 & 0
\end{bmatrix}$ and $B = \begin{bmatrix}
    0 & 0 \\  0 & 0 \\ \frac{3}{l_1^2(m_1+3 m_2)} & \frac{-3 (3 l_1+ 2 l_2)}{2l_1^2 l_2(m_1+3 m_2)} \\ 0 & \frac{3}{m_2 l_2^2}
\end{bmatrix}.$
This model is used by the agent for the controller design and the decision making.

\subsection{Free energy derivation} \label{app:FE_derivation}
The $F$ from (\ref{eqn:FE_components}) can be written as a sum of two terms ($V$ and $H$):
\begin{equation} \label{eqn:FE_VH}
    F = V+H = -\int q(U)\ln{p(\tau,U)} dU  + \int q(U) \ln{q(U)} d U.
\end{equation}
Here $q(U)  = \mathcal{N}(U:\mu^U,(\Pi^U)^{-1})$, $p(U)  = \mathcal{N}(U:\eta^U,(P^U)^{-1})$ and $ p(\tau/U) = \mathcal{N}(\tau:\tau^{g},(P^{\tau^g})^{-1})$, which can be written as:
\begin{equation} \label{eqn:rec_densities}
\begin{split}
    & q(U)   = \frac{1}{\sqrt{(2\pi )^r |{(\Pi^U)}^{-1}|}} e^{-\frac{1}{2} (U-\mu^U)^T \Pi^U (U-\mu^U) }    \\
    & p(U)  = \frac{1}{\sqrt{(2\pi )^r |(P^U)^{-1}|}} e^{-\frac{1}{2} (U-\eta^U)^T P^U (U-\eta^U) }    \\
    & p(\tau/U) =  \frac{1}{\sqrt{(2\pi )^n |(P^{\tau^g})^{-1}|}} e^{-\frac{1}{2} (\tau-\tau^g)^T P^{\tau^g} (\tau-\tau^g) } \\
\end{split}
\end{equation}
Considering a normalized recognition density $\int q(U)dU = 1$, and the fact that $\int q(U)(U-\mu^U)^T \Pi^U (U-\mu^U) dU = 1$, $H$ in (\ref{eqn:FE_VH}) reduces to $H = -\frac{1}{2} \ln{|\Pi^U|}$. The $V$ in (\ref{eqn:FE_VH}) can be simplified using the property $E = -\ln{p(\tau,U)}= -\ln{p(\tau/U)} -\ln{p(U)}$, and substituting (\ref{eqn:rec_densities}) in it.  $E$ simplifies to (after dropping constants):
\begin{equation} \label{eqn:E_expression}
    E = \frac{1}{2}(\tau - \tau^g)^T P^{\tau^g} (\tau-\tau^g) + \frac{1}{2} (U-\eta^U)^T P^U (U-\eta^U) 
\end{equation}

The relation between $V$ and $E$ can be expressed as $V(\tau,U) = \int dU q(U) E(\tau,U)$.
Here, $E(\tau,U)$ can be approximated through the Taylor series expansion around the mean $\mu^U$. So,  $ V(\tau,U)  = \int dU q(U) \big[ E(\tau,\mu^U) + \frac{ \partial E(\tau,\mu^U)}{\partial U}(U - \mu^U) + \frac{1}{2}(U - \mu^U)^T \frac{ \partial^2 E(\tau,\mu^U)}{\partial U^2}(U - \mu^U) \big] $. Since $\frac{ \partial E(\tau,\mu^U)}{\partial U} = 0$ for free energy minimization, the second term drops to 0. Since  $\int q(U)dU = 1$, $ V(\tau,U)  =  E(\tau,\mu^U) + \frac{1}{2} \int dU q(U) (U - \mu^U)^T \frac{ \partial^2 E(\tau,\mu^U)}{\partial U^2}(U - \mu^U)  $, which reduces to $ V(\tau,U)  =  E(\tau,\mu^U) +  \frac{1}{2} tr \Big( (\Pi^U)^{-1}  \frac{ \partial^2  E(\tau,\mu^U)}{\partial U^2} \Big)  $. Using (\ref{eqn:E_expression}) with $\eta^U = 0$, $V$ can be written as $    V = \frac{1}{2}(\tau - \tau^g)^T P^{\tau^g} (\tau-\tau^g) + \frac{1}{2} U^T P^U U + W$,  where $W =\frac{1}{2} tr \Big( (\Pi^U)^{-1}  \frac{ \partial^2  E(\tau,\mu^U)}{\partial U^2} \Big) $. Substituting it in (\ref{eqn:FE_VH}) after dropping constants gives:
\begin{equation} \label{eqn:F_with_W}
     F =  \frac{1}{2}(\tau^j -\tau^{j^g})^T P^{\tau^{j^g}} (\tau^j -\tau^{j^g})  + \frac{1}{2}U^T P^{U} U  -  \frac{1}{2} \ln{|\Pi^U|} + W.
\end{equation}
Since most of the FEP literature neglects $W$ in the free energy term, we will drop it for calculating $F$. However, we will keep $W$ in $F$ in the next appendix. Refer \cite{anil2021dynamic,friston2008variational} for the details of a similar mathematical procedure (variational inference) to evaluate $F$.

\subsection{Control confidence} \label{app:control_conf}
According to FEP, the free energy gradient is zero at optimal estimates. Therefore, the free energy gradient with respect to the covariance of action ($\Sigma^U = (\Pi^U)^{-1})$ is zero for optimal action $U$, i.e, $\frac{\partial F}{\partial \Sigma^U}$ = 0. Differentiating (\ref{eqn:F_with_W}) with $\Sigma^U$ and equating it to zero yields:
\begin{equation}
    \frac{\partial F}{\partial \Sigma^U} = \frac{1}{2}\Pi^U - \frac{1}{2}   \frac{ \partial^2  E}{\partial U^2} = 0, \implies \Pi^U =  \frac{ \partial^2  E}{\partial U^2}.
\end{equation}
Since $F$ in (\ref{eqn:FE_components}) has the same components of $U$ as that of $E$ in (\ref{eqn:E_expression}), we abuse the notation to write $ \Pi^U =  \frac{ \partial^2  F}{\partial U^2}$.

\subsection{Task specific controllers} \label{sec:task_controller}
In this work, three variations of active inference controllers are introduced for three tasks: i) position control for task 1, ii) velocity control for task 2, and iii) acceleration control for task 3. 

\subsubsection{Task 1 (position control)} \label{sec:task1}
Task 1 involves controlling the robot arm to move to a goal configuration $\theta^g$ with zero goal velocity and acceleration. The controller uses all four terms of (\ref{eqn:U_dot_deriv}) with $\Dot{\theta}^g = \Ddot{\theta}^g = \eta^U = [0 \ 0]^T$ as:
\begin{equation}
\begin{split}        
     \Dot{U}  = - \frac{\partial X}{\partial U}^T  \Big[ \begin{smallmatrix}
        P^{\theta^g} & 0 \\ 0 & P^{\Dot{\theta}^g} 
    \end{smallmatrix}  \Big] \Big[ \begin{smallmatrix}
    \theta - \theta^g & 0 \\ 0 & \Dot{\theta} 
    \end{smallmatrix} \Big] -  \frac{\partial  \Ddot{\theta}}{\partial U}^T P^{\Ddot{\theta}^g} \Ddot{\theta} - P^U U 
\end{split}
\end{equation}
The goal is to find the unknowns $\frac{\partial X}{\partial U}$ and $\frac{\partial  \Ddot{\theta}}{\partial U} $. We assume a zero jerk motion for the robot using $\frac{\partial  \Ddot{\theta}}{\partial U}  = \big[ \begin{smallmatrix}
    1 & 0  \\ 0 & 1 
\end{smallmatrix} \big]$ and $\frac{\partial  \Dot{X}}{\partial U}  = \Big[ \begin{smallmatrix}
    1 & 0 \\ 0 & 1 \\ 1 & 0 \\  0 & 1 
\end{smallmatrix} \Big]$ . Differentiating (\ref{eqn:LTI}) with respect to $U$ gives the remaining unknown in the controller design as $\frac{\partial X}{\partial U} = A^{-1} (\frac{\partial  \Dot{X}}{\partial U}  - B)$. 

\subsubsection{Task 2 (velocity control)}
Task 2 involves controlling the robot arm to maintain a goal velocity $\dot{\theta}^g$ with zero acceleration. The controller uses the last three terms of (\ref{eqn:U_dot_deriv}) with $\Ddot{\theta}^g = \eta^U = [0 \ 0]^T$. The goal is to find the unknowns $\frac{\partial \Dot{\theta}}{\partial U}$ and $\frac{\partial \Ddot{\theta}}{\partial U}$. Differentiating \ref{eqn:LTI} with $U$, and assuming $   \frac{\partial \Dot{X}}{\partial U} = O$ gives $\frac{\partial \Dot{\theta}}{\partial U} = -[    I_{2 } \ O_{2 }]A^{-1}B$. Assuming jerk free motion gives the last unknown in the update equation as $\frac{\partial  \Ddot{\theta}}{\partial U}  = \big[ \begin{smallmatrix}
    1 & 0  \\ 0 & 1 
\end{smallmatrix} \big]$.

\subsubsection{Task 3 (acceleration control)}
Task 3 involves controlling the robot arm to maintain a goal acceleration $\ddot{\theta}^g$. The controller uses the last two terms of (\ref{eqn:U_dot_deriv}) with $\eta^U = [0 \ 0]^T$. The goal is to find the unknown $\frac{\partial \Ddot{\theta}}{\partial U}$. Differentiating (\ref{eqn:LTI}) with $t$, and substituting (\ref{eqn:LTI}) in it yields $\Ddot{X} = A^2 (X-X^e)+ABU+B\Dot{U}$. Differentiating it with $U$ and using $\frac{\partial \Dot{U}}{\partial U} = 0$ gives $\frac{\partial \Ddot{X}}{\partial U} = A^2 \frac{\partial X}{\partial U} + AB$. From Section \ref{sec:task1} we know  $\frac{\partial X}{\partial U} = A^{-1} (\frac{\partial  \Dot{X}}{\partial U}  - B)$ and $\frac{\partial  \Dot{X}}{\partial U}  = \Big[ \begin{smallmatrix}
    1 & 0 \\ 0 & 1 \\ 1 & 0 \\  0 & 1 
\end{smallmatrix} \Big]$. Therefore, $\frac{\partial \Ddot{\theta}}{\partial U} = [I_2 \ O_2] \frac{\partial \Ddot{X}}{\partial U} = [I_2 \ O_2] A^2 (A^{-1}(\frac{\partial  \Dot{X}}{\partial U}  
 - B) + A B)$.

\addtolength{\textheight}{-12cm}   




\end{document}